# Robustness Analysis of Deep Learning Models for Population Synthesis

Daniel Opoku Mensah[a], Godwin Badu-Marfo[a], Bilal Farooq[a,*]

[a]*Laboratory of Innovations in Transportation (LiTrans), Toronto Metropolitan University, Toronto, Canada*

**Abstract**

Deep generative models have become useful for synthetic data generation, particularly population synthesis. The models implicitly learn the probability distribution of a dataset and can draw samples from a distribution. Several models have been proposed, but their performance is only tested on a single cross-sectional sample. The implementation of population synthesis on single datasets is seen as a drawback that needs further studies to explore the robustness of the models on multiple datasets. While comparing with the real data can increase trust and interpretability of the models, techniques to evaluate deep generative models' robustness for population synthesis remain underexplored. In this study, we present bootstrap confidence interval for the deep generative models, an approach that computes efficient confidence intervals for mean errors predictions to evaluate the robustness of the models to multiple datasets. Specifically, we adopt the tabular-based Composite Travel Generative Adversarial Network (CTGAN) and Variational Autoencoder (VAE), to estimate the distribution of the population, by generating agents that have tabular data using several samples over time from the same study area. The models are implemented on multiple travel diaries of Montréal Origin-Destination Survey of 2008, 2013, and 2018 and compare the predictive performance under varying sample sizes from multiple surveys. Results show that the predictive errors of CTGAN have narrower confidence intervals indicating its robustness to multiple datasets of the varying sample sizes when compared to VAE. Again, the evaluation of model robustness against varying sample size shows a minimal decrease in model performance with decrease in sample size. This study directly supports agent-based modelling by enabling finer synthetic generation of populations in a reliable environment.

## 1. Introduction

Individuals with their locations and sociodemographic attributes, are needed for the development of agent-based travel demand micro-simulation. It is possible to accomplish this by creating synthetic data from real sample data, while preserving the statistical features. Population synthesis is a method that aims to anticipate populations in social and geographic areas under a variety of conditions that often reflect future population profiling (Mu¨ller, 2017; Borysov et al., 2019). As the focus on agent-based modelling in the transportation sector has increased, this topic has recently attracted more attention. In activity-based travel demand forecasting, population synthesis refers to the expansion of a representative sample of persons with socioeconomic and spatio-temporal characteristics into a broader population under study. To address topics like fairness, household composition, social coherency, and ageing in more detail, transportation modelers look into the Liu and Tuzel (2016) distributional effects in the socio-economic and geographic realm (Borysov et al., 2019).

---

* Corresponding author. Tel.: +1-416-979-5000.
  *E-mail address:* bilal.farooq@ryerson.ca

The use of phone surveys, print or online surveys, trip diaries for households or individuals, and other methods has been employed by Census organisations over the years to gather data for transportation modelling. Having complete access to individual data provides utmost advantages for modelling and evaluating varied individual behaviour and is extremely beneficial when assisting decision-making (Namazi-Rad et al., 2017). The public's concern over how their data is used and shared and the tightening of privacy regulations have made it difficult to obtain access to personal data for analysis. Rather effectively and at scale, data synthesis can provide the analyst with genuine data to work with while preventing the data from being viewed as identifiable personal data (El Emam et al., 2020). Therefore, synthesizing the population for future years based on the few qualities predicted by planners, such as the population per zone, possibly income, and hopefully car ownership, is a suitable course (de Dios Ortúzar and Willumsen, 2011).Aside from the legitimate privacy concerns linked to access to personal data, it is also expensive to gather data for transportation modelling via paper or online surveys, personal trip diaries, and phone polls because these techniques require much time and work (Ghorpade, 2018). The costs multiply when carrying out these trip surveys with longer observation periods. Therefore, the easiest way to address concerns about cost and privacy is through the synthetic production of agent data.

Utilizing population synthesis methods like Iterative Proportional Fitting (IPF), Combinatorial Optimization (CO), and Markov Chain Monte Carlo (MCMC) simulation, researchers reconstruct representative individuals of a population using sample data, and other information (Badu-Marfo et al., 2022). In early studies on population synthesis, these techniques were employed to create a general-purpose population by fitting the joint distributions of several variables to their sampling distributions (Zhuo, 2017). These techniques are used in travel demand forecasting to generate a population for assessing travel activities by taking into account people's travel patterns and related socio-economic and demographic factors (Zhuo, 2017).Recently, deep generative models have been introduced to find patterns in input data and subsequently produce new data observations that are remarkably similar to the original dataset Kieu et al. (2022). Some well-known deep generative models are Boltzmann machines, Variational Autoencoders, and Generative Adversarial Networks (GANs).

Recent studies have expanded the models from unsupervised to semi-supervised environments though only for classification problems (Wan et al., 2017). These models offer a robust statistical framework for assessing uncertainty and drawing inferences from big data sets when noise, sparsity, and bias are present (Lopez et al., 2020). Deep generative models (Goodfellow et al., 2014) have opened a path toward large-scale problems to deal with model performance on different datasets. Deep generative model approaches that have been used to estimate the underlying joint distribution of a population using variety of travel surveys have shown consistent and accurate generation of synthetic populations. The features that are generated from deep generative models are important for many agent- based model systems for which the ability to carry out long-term forecasts of detailed populations is essential Garrido et al. (2020). Hence, there is the need to ensure robustness of the models in generating synthetic data. In this study, we define robustness as the degree to which a model's performance changes when using new data. Ideally, it is expected that model's performance deviate significantly.

In practice, most of the existing studies have implemented population synthesis based on deep generative models on single datasets which is one dimensional. This drawback is seen as a research interest that needs further studies to develop models that are robust and efficient when evaluated on multiple travel diaries. We investigate how deep generative models are robust for population synthesis. Robust deep generative models can improve prediction performance and learning stability of imbalanced datasets. How a model responds to unseen and imbalanced data beyond available train and test datasets raises questions. The robustness of a model to changes in a dataset can determine its practicality for real-world applications. Wang (Wang et al., 2020) proposes deep generative classifier modelled by latent variable where the latent variable capture the direct cause of target variable. To effectively handle imbalance data, they design a deep generative model to simultaneously mine the direct cause of target label and build a stable generative imbalanced classifier. The proposed model mitigates unstable prediction and low performance that arises from imbalanced dataset.

In this paper, we develop a method for evaluating the robustness of deep generative models to changes in datasets from multiple travel diary surveys. Specifically, we adopt the tabular component of the Composite Travel Generative Adversarial Network (CTGAN), a novel deep generative model proposed by (Badu-Marfo et al., 2022) and Variational Autoencoder (VAE) and evaluate the effectiveness of the models on variety of datasets. Our main contributions of this paper are summarized as follows:

- First, we connect the area of population synthesis with deep generative modeling and demonstrate how the models estimate the underlying joint distribution of a population capable of reconstructing synthetic agents having tabular (e.g., age, sex, employment status and permit) data of a travel survey. In doing so, we investigate the socio-demographic and economic attributes to create a synthetic population with statistical significance from multiple surveys.
- Second, we determine the robustness of the two deep generative models, CTGAN and VAE, for the generation of synthetic populations and demonstrate that their performance is affected when used on a variety of travel surveys of the same area. This is done by performing bootstrapping and repeating the experiments using selected samples of the datasets in order to estimate the confidence intervals of the error metrics to determine how the model will perform on unseen data.
- Finally, we evaluate the performance and similarity of synthesized tabular data distributions of CTGAN with VAE by comparing the statistical results of attributes for tabular samples of travel data from the Montreal Origin-Destination (OD) survey of 2008, 2013, and 2018.

To the best of our knowledge, this is the first time such systematic robustness analysis has been performed. The rest of the paper is organized as follows. In Section 2, we briefly review existing population synthesis approaches and discuss how deep generative models are robust to population synthesis. In Section 3, we describe the methodology and present how the models perform robustness. In Section 4, a case study and evaluation procedure are presented. Finally, in Section 5, we present the results and offer a discussion with future directions.

## 2. Literature Review

Population synthesis aims to effectively and efficiently use the available micro samples—along with the complementary aggregated information on each attribute of interest—to produce a realization of the population that could satisfy the underlying population structure as much as possible (Sun and Erath, 2015). Different kinds of population synthesis, including those for households, people, and dwelling populations, are developed to meet diverse needs. Creating a synthetic model of the real population that is statistically representative has been proposed using a variety of techniques. The two main approaches for accomplishing population synthesis include the conventional approach and the deep generative modelling approach. The conventional population synthesis process usually begins by determining the socio-demographic features required of the synthesized households and individuals (Guo and Bhat, 2007). Similarly, the deep generative modelling technique needs generative models that capture the inner probability distribution that creates a class of data to generate related data (Oussidi and Elhassouny, 2018).

The conventional methods for population synthesis include Markov-based, and fitness-based synthesis (FBS), combinatorial optimization, iterative proportional fitting (IPF), iterative proportional updating (IPU), and other cutting-edge techniques (Ramadan and Sisiopiku, 2019). The IPF technique, developed by Beckman et al. (1996) is a mechanism for generating a synthetic baseline population of individuals and families for microscopic activity-based models. They used the IPF technique to create limited maximum entropy estimations of the true proportions in the population multiway table. Ye et al. (2009) created the iterative proportional updating approach, a heuristic method, to remedy the shortcomings of the IPF methodology. Specifically, the IPU approach tackles the issue of control for individual-level attributes and joint distributions of personal traits. Abraham et al. (2012) created the combinatorial optimization strategy, which is a flexible approach that can match targets for both household- and individual-level features at many agent levels. In general, a combinatorial optimization strategy is easier and more straightforward than IPF. Markov process-based methods were created to overcome such problems and provide a technique that actually synthesizes populations rather than cloning them. Farooq et al. (2013), who devised a Markov chain Monte Carlo (MCMC) simulation-based approach for synthesizing populations, made the first significant attempt in this field in 2013. A dependent sequence of random draws from challenging stochastic models can be simulated using the suggested strategy, which is a computer-based simulation technique. The fitness-based synthesis approach, proposed by Ma and Srinivasan (2015), directly provides a list of households that match numerous multilevel controls without the requirement to determine a combined multiway distribution, addressing the IPF approach's incapacity to cope with multilevel controls. However, while simulation-based techniques outperform IPF techniques, these methods still have limitations in the context of synthetic population generation (Lederrey et al., 2022).The main issue is that the models are working with conditionals only. This can be an advantage if only this information is available. Later, in Sun and Erath (2015), a Bayesian network approach for population synthesis was proposed. The idea is to use a Bayesian network model to estimate a probabilistic graph of the data-generation process. Sun et al. (2018) proposed a hierarchical mixture modelling framework for population synthesis that extends their previous work using Bayesian networks.

The evaluation of model robustness for population synthesis has been studied further in the conventional approaches. For instance, in their research on IPF technique, Beckman et al. (1996) evaluated the model's robustness and observed that the synthesis error was under 5% for 19 homogenous zones and under 15% for 3 heterogeneous zones. The tolerable error for heterogeneous zones reveals the IPF robustness. Similarly, Lim (2020) evaluated the robustness of IPU for population synthesis based on Australian census data by conducting multiple experiments to test the efficacy and robustness of the IPU algorithm. The experimental results reveal that the model was an effective and efficient process for generating synthetic populations. Sun and Erath (2015) also demonstrated the performance of the proposed Bayesian network approach by conducting numerical experiments on generating synthetic population in Singapore. The population data used in the following experiments comes from Household Interview Travel Survey in 2012 (HITS2012), which is conducted by the Land Transport Authority of Singapore. To evaluate robustness and solve model complexity that avoids overfitting, the model introduced penalty on size of parameters. While the conventional approaches demonstrate robustness for population synthesis, the studies are implemented on single datasets which is a drawback that needs to be further studied to determine the models' response to changes in variety of datasets. This is because how robust a model changes in a dataset can determine how practical it can be for real-world applications.

Deep generative models have had a recent upsurge in the machine learning literature (Goodfellow et al., 2016). These models are extensions of the probabilistic graphical models but with the use of neural networks as a means to provide more flexible models (Garrido et al., 2020). For the past decade, deep unsupervised learning has been dominated by generative models. This is due to the fact that they provide a very effective method of understanding and analyzing unlabeled data. The goal of generative models is to reproduce a class of data by capturing the internal probability distribution that produces that class of data (Oussidi and Elhassouny, 2018). Although deep generative models have a similar focus on high-quality synthesis, their approaches differ greatly (Regenwetter et al., 2022). Generally, the models can often be separated into two groups. This includes energy-based models, where the joint probability is determined using an energy function, such as Boltzmann machines and their variants, deep belief networks, as well as cost function-based models (autoencoders and generative adversarial networks) (Oussidi and Elhassouny, 2018). Boltzmann Machines are an energy-based model created by Fahlman et al. (1983) that aims to learn and identify an energy function that correlates smaller values with the suitable configurations and greater values with the wrong ones, both inside and outside the training instances. The Binary Boltzmann Machine, Restricted Boltzmann Machine, and Deep Boltzmann Machine are some examples of certain types of Boltzmann machines. Deep belief networks, another deep architecture with numerous hidden layers, revolutionized deep learning when they were originally introduced in 2006 (Oussidi and Elhassouny, 2018).

Autoencoders are unsupervised embedding algorithms that combine a decoder that reconstructs the design as precisely as feasible from the latent space with an encoder that maps an input design into a (usually) lower-dimensional latent space. Variational autoencoders (VAE), a common type of autoencoder, make the assumption that the coding layer has a mean and variance that follow a Gaussian distribution Oussidi and Elhassouny (2018). VAE (Kingma and Welling, 2013) focuses on minimizing the dimensions of the data into an encoded vector in the latent space. Due to its reduced dimensionality, data can then be generated more easily in this latent space. For instance, Borysov et al. (2019) created a synthetic population using a VAE. They showed that for producing complex data, the VAE model outper-formed both IPF and Gibbs sampling. GANs, on the other hand, are used to estimate generative models through an adversarial process in which we simultaneously train two models Goodfellow et al. (2014). It has a generative model, *G* that captures the distribution of the training data and a discriminative model, *D* that estimates the likelihood that a sample came from the training data rather than *G*. GANs and VAEs are two specific types of DGMs that have successfully combined pictures, text, and tabular data in a variety of fields (Regenwetter et al., 2022). Generative adversarial networks proposed by Goodfellow et al. (2014) have become predominant for deep generative modelling because of their ability to generate synthetic data for a specified domain that are different and hardly unidentifiable from other real data. The authors proposed an adversarial technique for predicting generative models by jointly training two models: a generative model *G* that captures the data distribution and a discriminative model *D* that estimates the probability that a sample came from the training data rather than *G*. The model utilizes the zero-sum non-cooperative game idea, in which *G* and *D* are trained to play against each other until they establish a Nash equilibrium.

During model learning, they train the generator and discriminator concurrently. The generator generates a batch of samples, which, along with real samples from the domain, are fed into the discriminator and categorised as real or fake. The discriminator is then updated in the next round to improve its ability to discriminate between real and false examples, and the generator is estimated based on how far the generated samples deceived the discriminator. While the primary application of GANs has been the generation of image data, with a particular focus on human faces Alqahtani et al. (2021), researchers have also developed specific architectures for tabular data. Thus, it allowed researchers to switch their focus to more general synthetic tabular data generation rather than synthetic population generation. Researchers have also developed their own GAN structures to generate synthetic populations in the transportation community. For example, Garrido et al. (2020) develop their own GAN structure based on the WGAN to use tabular data. They show that this new model was statistically better than IPF techniques, Gibbs sampling, and the VAE of (Borysov et al., 2019). Finally, Badu-Marfo et al. (2020) created a new GAN named Composite Travel GAN (CTGAN). Their GAN is based on the Coupled GAN (CoGAN) Liu and Tuzel (2016) and is used to generate the table of attributes for the population and the sequence of Origin-Destination segments. They show that the CTGAN outperforms the VAE statistically.

The deep generative models can model complex real-world data and that gives it enough attention in the research community. The models are used to synthesize data while restoring balance in imbalance data due to their concept of adversarial learning. In population synthesis-based deep learning models, the performance of synthesized populations is evaluated mostly on single datasets. Badu-Marfo et al. (2020) proposed Composite Travel Generative Adversarial Network (CTGAN), a deep generative model and evaluated the performance of the synthesized outputs based on distribution similarity, multi-variate correlations and spatio-temporal metrics on Montreal OD Survey 2008. Similarly, to predict rare feature combinations in population synthesis, Garrido et al. (2020) proposed Wasserstein Generative Adversarial Network (WGAN) and the Variational Autoencoder (VAE), and adopt these algorithms for a large-scale population synthesis application. The models are implemented on a Danish travel survey with a feature-space of more than 60 variables. The models make predictions for data inputs that are drawn from the same distributions and generate synthetic data for which the deep learning model maybe highly confident of making robust prediction yet wrong on different datasets. While previous studies have evaluated the performance of deep generative models on sin- gle datasets, there has not to the best of our knowledge any comprehensive evaluation of the models' robustness under variety of datasets. Indeed, we propose an effective evaluation of the models' performance under different datasets. The reason for this is that the evaluation would demonstrate whether the model performs well in the deployment environment, which often differs from the environment in which training data was gathered (Quinonero-Candela et al., 2022). Additionally, it should test whether the model will perform well across all relevant subpopulations and whether performance will deteriorate in unexpected ways as the deployment environment evolves over time.

## 3. Methodology

In this section, we describe the robust deep generative models for population synthesis. We adopt the tabular-based component of Composite Travel Generative Adversarial Networks (CTGANs) (Badu-Marfo et al., 2020) and tabular-based VAE for synthesizing the tabular attributes.

## 3.1. Composite Travel Generative Adversarial Network (CTGAN)

The Composite Travel Generative Adversarial Networks (CTGAN) is built for training the joint distribution of tabular travel features and sequential trip chain locations of an individual in a simultaneous manner, taking inspiration from the CoGAN proposed by Liu and Tuzel (2016).

The CoGAN model is designed to learn a joint distribution of multi-domain images from data. According to the authors, the model is used for a variety of joint distribution learning tasks, such as generating a joint distribution of colour and depth images and training a joint distribution of facial images with various features. The CTGAN model is made up of two GAN networks: the Tabular model and the Sequence model. The purpose of the tabular component is to learn the joint distribution of the basic socio-demographic attributes in the travel diary while the sequential component learns the distribution of the trips taken by an individual per day. The model is designed in such a way that each component is implemented as an independent network that trains its parameters depending on the fundamental data distribution. The model then combines pairs of tabular features with sequential locations of an individual in a population. In this study, we adopt only the tabular component of the CTGAN model and trains the parameters based on the socio-demographic and economic attributes of the Montreal OD survey data of 2008, 2013 and 2018.

In this paper, we modify the GAN model by integrating a multi-layer perceptron (MLP) network to enable the generation of tabular data for the Montreal OD travel survey data. The Tabular CTGAN architecture is shown in Fig. 1. The objective of the Tabular component in the CTGAN is to generate the table of records on an agent's socio-demographic and economic attributes (i.e. Age, Sex, Status, Gender, Permit) which can be in numerical as well as categorical types. The model is able to synthesize both types of tabular attributes. The model is made up of an independent GANs architecture with a single generator ($GT$) and discriminator ($DT$). The generator network generates new sample data that are identical to the real data. The generator has an input layer that receives random noise drawn from a Gaussian distribution with a dimension size equal to the size of the real data. The discriminator model determines whether the data is real or fake. The discriminator takes input from both the generator and the real samples and computes the loss function.

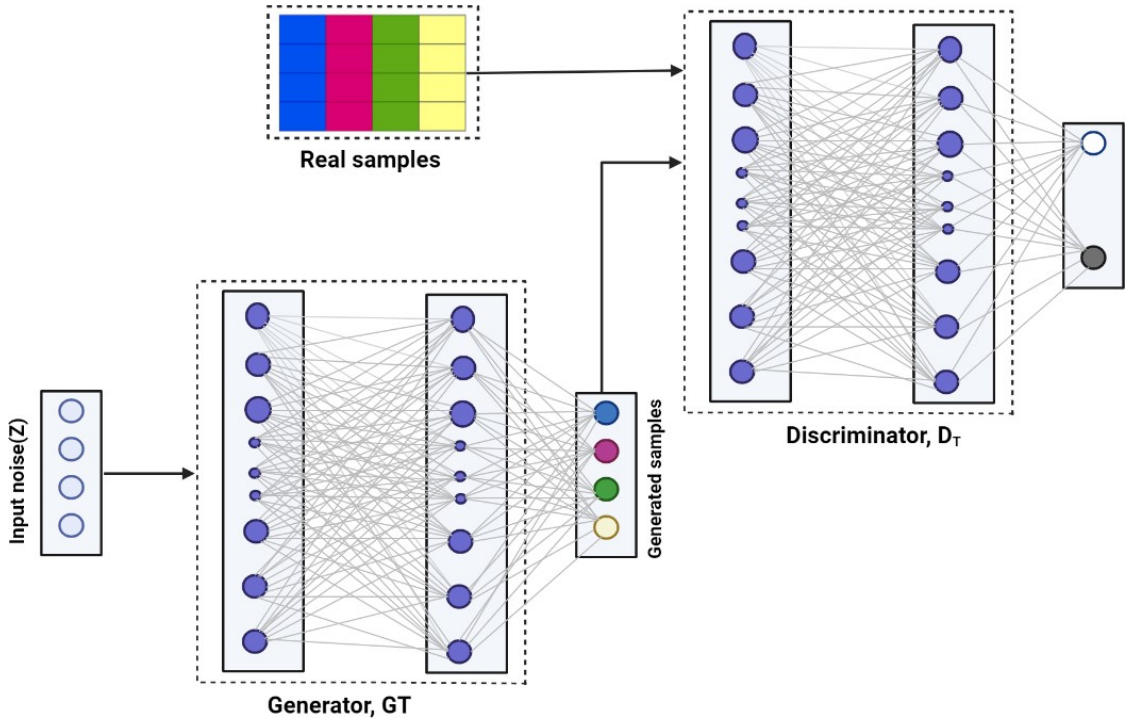

Fig. 1. The structure of the Tabular component of CTGAN

During GAN training, the model parameters of the discriminator are fixed when training the generator while the model parameters of the generator are fixed when training of the discriminator. The CTGAN model's network layer is made up of a Multi-Layer Perceptron (MLP) neural network with neurons from each layer coupled to neurons from the next layer. The Multi-Layer Perceptron (MLP) network is used in both the generator and discriminator models. MLP neural networks are made up of units stacked in layers that are made up of nodes (Delashmit et al., 2005). In our study, we explore fully linked networks which have nodes that connect to every node in successive layers. Each MLP is made up of three layers: an input layer, one or more hidden layers, and an output layer. The input layer sends the inputs to the subsequent layers within the hidden units. In addition to the weights, the nodes from each of the hidden and output units have thresholds connected with them. The activation functions of the hidden unit nodes are nonlinear, whilst the activation functions of the outputs are linear. We use multiple hidden layers in this network due to the depth of features learnt in the neural network. Each layer has a bias and its output is activated via a Rectified Linear Unit (ReLU). The category vectors utilize a non-linear activation function called softmax, whereas the numerical vectors use a linear activation function. Eventually, as the generator model's final output, we combine the output layers. Eventually, the final output of the generator model is obtained through a combination of the output layers.

### 3.2. Variational Autoencoder (VAE)

The Variational Auto-Encoder (VAE) was proposed by Kingma and Welling (2013), as an alternate deep learning approach to estimate a population distribution into a compressed lower dimensional latent space using a neural network called the "encoder" that is supported by an auxiliary neural network named the "decoder", acting as a generator by drawing random samples from the distribution of the latent space. The method is particularly interesting from the perspective of population synthesis as it enables a rendering of agents that are on the one hand diverse, but yet, from a statistical point of view, similar to the agents in the sample data (Garrido et al., 2020). The Tabular VAE model is shown in Fig. 2.

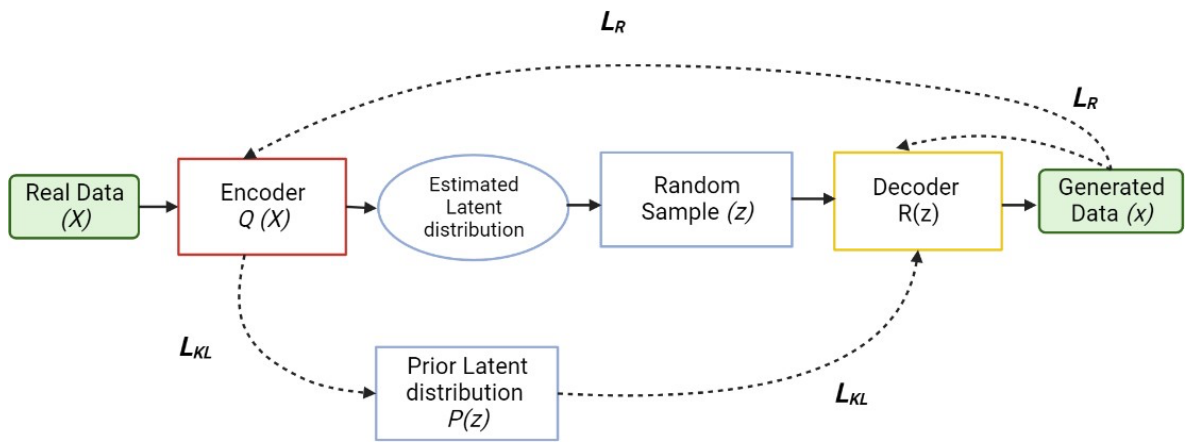

Fig. 2. The structure of the Tabular VAE model

Consider the Montreal OD Survey dataset on mobility of the population agents which is $X = (X_1, X_2, X_3, ...X_n)$ where n is the number of attributes. The encoder $Q$ maps the input data ($X$) into the parameters of latent prior distribution $P(z)$ e.g., a vector of the mean and standard deviation of multivariate normal distribution), whereas the decoder R generates the data ($x$) mimicking $X$ from the $z$ sampled from $P(z)$. The decoder parameter $\theta_d$ estimated together with the encoder parameter $\theta_e$ based on a reparameterization trick proposed by Kingma et al. (2019). The VAE is trained to minimize the discrepancy between the input and generated data. The model utilizes the VAE loss function. The loss function of VAE is sum of the loss function of the real data $L_R$ and Kullback-Leibler (KL) divergence $L_{K_L}$. KL divergence measures the divergence between two probability distributions.

*3.3. Problem Setting*

In a region under consideration at any point in time there exists a true population, we assume that the Montréal OD Survey dataset on mobility of the population agents (i.e. households, families or individuals) are characterized by a set of basic attributes $X = (X_1, X_2, X_3, ...X_n)$ where n is the number of attributes. These attributes may be discrete (e.g. sex, status, permit) or continuous (e.g. age). In this study, we employ the deep generative models to predict the joint distribution of a true population using sample partial views with tabular attributes, from which we can simultaneously construct synthetic agents with tabular features. First, we preprocess the datasets to ensure that the output conditions for the models are fulfilled, as detailed in the preceding section. The Tabular CTGAN model employs an MLP neural network, which is made up of four layers of fully connected network layers. The first hidden layer has an arbitrarily defined size of 100 neurons, preceded by 50 neurons in the last hidden layer. The neuron sizes were randomly selected, with the optimum option based on the model's training performance and the distribution of the output layer.

During the model training, we employ the gaussian distribution to produce random noise data with the shape (1000), and that will be used as the input for the generator network. The number of neurons from the first to the fourth layers of the entire connection layer is hidden1, hidden2, hidden3 and hidden4. Leaky Rectified Linear Unit (Leaky RELU) is adopted as the activation function for the hidden layers. Then we apply a Sigmoid activation to the output layer for the numeric variable and Softmax activation to the last hidden layer for categorical variables. Similarly, the Tabular VAE model employs the MLP neural network which is made up of three layers of fully connected layers both for the encoder and decoder models. The first hidden layer has an arbitrarily defined size of 200 neurons, preceded by 100 neurons and 50 neurons in the last hidden layer. Rectified Linear Unit (RELU) is used as the activation function for the hidden layers while softmax is used as activation functions for the output layers.

*3.4. Robust Evaluation of Confidence Intervals with Bootstrapping Method*

The development of good generative models depends on accurate performance evaluation of the models which is constrained by limited data and sampling biases. Hence, the adoption of confidence interval offers an opportunity into determining the uncertainty of the error and model performance. With the confidence interval, we can compute the upper and lower bound around the estimated value of the errors while the actual error remains inside or outside the bounds. In this study, we utilize the bootstrap method for the confidence intervals in order to critically and systematically evaluate the robustness of the models trained at varying sample sizes when synthesizing individual level attributes or populations. The bootstrap method is a resampling technique for estimating a sampling distribution Raschka (2018), and in the context of this study, we are particularly interested in estimating the uncertainty of a performance estimate – the error. In brief, the idea of the bootstrap method is to generate new data from a population by repeated sampling from the original dataset with replacement.

During bootstrapping, we split the entire dataset into two parts, a training, and an independent test set. We use the k-fold cross-validation method for model selection with an idea of keeping an independent test dataset. We withhold the test dataset during training and model selection, to avoid the leaking them in the training stage. Fig. 3 shows the model architecture of the bootstrapping method designed to achieve the confidence intervals. The process for the bootstrapping method is as follows:

- Given a dataset $(x_1, x_2, x_3, ... x_n)$ of size *n*.
- For *b* bootstrap rounds draw one single instance from this dataset and assign it to the $j_t h$ bootstrap sample.
- Repeat step 2 until the bootstrap sample has size *n* – the size of the original dataset. Each time, draw samples from the same original dataset such that certain examples may appear more than once in a bootstrap sample and some not at all.
- Fit the deep generative model to each of the b bootstrap samples and compute the error. Use out-of-bag samples as test sets for evaluating the model on the training data.
- Compute the mean error of the error estimates over the b error estimates

Let us assume that a sample that has been drawn from a normal distribution, we can compute the 95% confidence interval of the bootstrap estimate starting with the mean of the error metric, herein MSE.

$$MSE = \frac{1}{b} \sum_{i=1}^{b} MSE_i \qquad (1)$$

We obtain the confidence interval by using a more robust and computationally straight-forward approach which is the percentile method which uses the lower and upper confidence bounds.

In this study, we use two tabular-based deep generative models, Tabular CTGAN and Tabular VAE to efficiently generate synthetic samples from multiple travel diaries of Montreal OD Survey of 2008, 2013 and 2018. In this approach, random samples are selected from the original sample with sizes of 25%, 50% and 75%. The varying selected samples and the original samples are independently trained as inputs to each of the models. We generate synthetic data using the proposed deep generative models and obtain results for each model trained at the varying selected samples. To show the robustness of our deep generative models, we perform bootstrapping on the train and test sets and repeat our experiments using the selected samples from the train sets. For each model, we set random seeds on different sample sizes of the training and test set. Then, we compute the confidence intervals from the different random seeds. Thus, the results of the models are obtained by generating multiple synthesized sets under different bootstraps and random seeds.

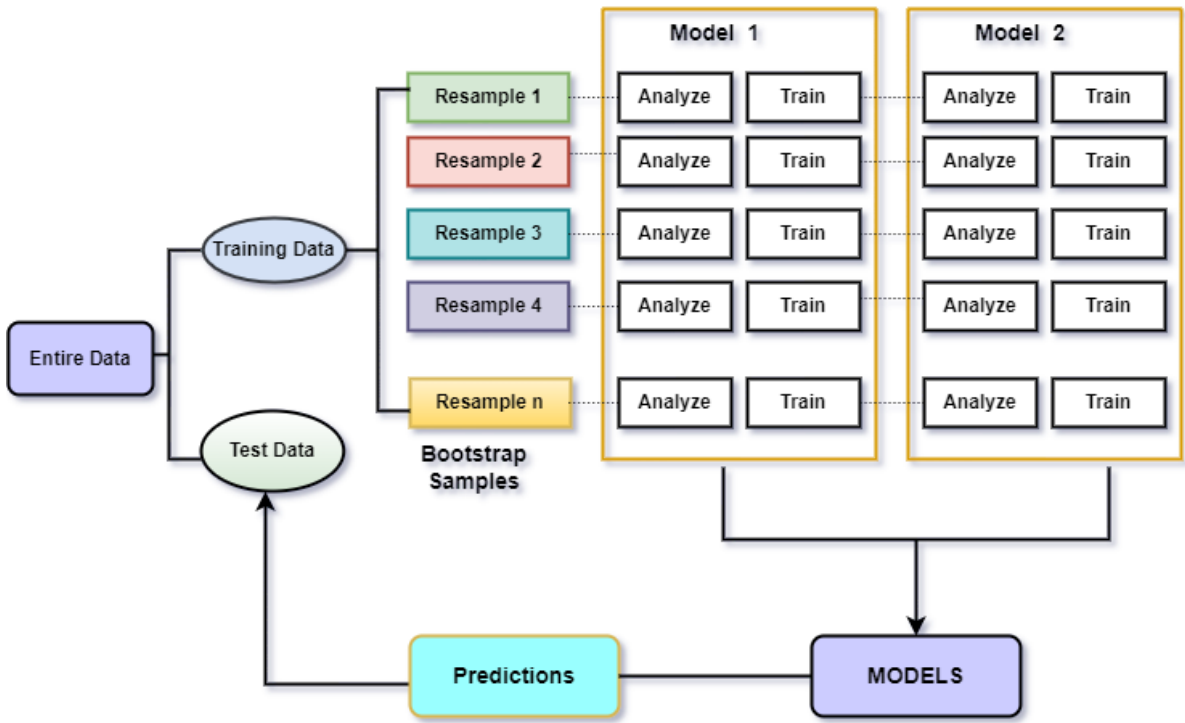

Fig. 3. Model Architecture of Bootstrapping Method

## 4. Case Study and Data

Greater Montreal Area is used as the case study for this work.

*4.1. Data*

The proposed CTGAN model is used in a real-world case study to generate a synthetic population for survey travel data from the Montreal Origin-Destination (OD) survey (Origine-Destination, 2008). Since 1970, the survey has been conducted every five years, and it is one of the key sources to support transportation planning in Quebec (Hatzopoulou et al., 2013). It is a trustworthy and thorough source of information about how people travel around in the Greater Montreal area by foot, bicycle, bus, metro, rail, and vehicle. The survey is comprised of socio-demographic information at the individual and household levels, as well as a travel diary for each trip (i.e., trip origin, destination, purpose, mode of transportation). Telephone interviews were used to choose survey participants at random from the Montreal population. The interviews were conducted during a period when most urban travel habits were constant. In 2008, 66,100 households (representing 4% of the population) were interviewed including 156,700 individuals comprising 319,900 trips (Origine-Destination, 2008) while in 2013, 78,731 households were interviewed including 188,746 individuals comprising 410,800 trips. In 2018, 73,424 households were interviewed including 168,893 individuals comprising 358,056 trips. In this study, we used the 2008, 2013, and 21018 surveys and synthesize individual socio- economic and demographic variables such as age, sex, employment status and permit. Table 1 shows the selected attributes for the Montreal OD Survey.

Table 1. Description of Variables for Montreal OD Survey

| Column    | Type        | Description                       |
|-----------|-------------|-----------------------------------|
| *P_AGE*   | numeric     | Age of the respondent             |
| *P_SEX*   | binary      | Gender of the respondent          |
| *PERMIT*  | categorical | Driving permit type of respondent |
| *P_STATUT*| categorical | Employment status of respondent   |

*4.2. Data Preparation*

In this study, we deal with mixed data types for generating survey data poses two challenges: numeric representation, and reversibility. Since machine learning algorithms only take numerical values, there was the need to encode the categorical variables into numeric data using encoding techniques (Potdar et al., 2017). Both categorical and binary variables are indexed numerically and one-hot encoded while the numeric variables are scaled and normalised within a range of one (-1) and one (+1) (Badu-Marfo et al., 2022). The deep generative modelling pre-processing approach requires input data in the form of numbers. The idea behind the tabular input data is to be able to invert the output layer and make it readable in the same format as the input data. In this study, we employed the label encoding and one-hot encoders in Sci-kit learn library because of their ability to reverse the output layer format into a readable format similar to the input layer.

**5. Evaluation Metrics and Results**

This section discusses the evaluation metrics and the results obtained from evaluating the fitness of the synthesized population using similarity benchmarks on the statistical distribution. The experimental results are analyzed by comparing the two robust deep generative models and investigating their robustness to the various datasets.

*5.1. Evaluation Metrics*

To evaluate the predictive performance of each deep generative model, three metrics were considered: i) the Mean Absolute Error (MAE); ii) the Mean Squared Error (MSE); and, iii) the Root Mean Squared Error (RMSE). The MAE is the average of all absolute errors and it is computed as:

$$MAE = \sum_{i=1}^{n} |x_i - y_i| \qquad (2)$$

where $n$ is the number of errors, $|x_i - y_i|$ is the absolute error Similarly, the MSE which measures the prediction accuracy of a model by computing average magnitude of the error is computed as:

$$MSE = \frac{1}{n}\sum_{i=1}^{n} (|x_i - y_i|)^2 \qquad (3)$$

The RMSE is the square root of the mean squared error (MSE). As a metric for normally distributed errors, RMSE is computed as:

$$RMSE = \sqrt{\frac{1}{n}\sum_{i=1}^{n} (|x_i - y_i|)^2} \qquad (4)$$

*5.2. Evaluation Results*

In this section, we discuss the results achieved from the implementation of the models. We perform sensitivity analysis by critically analyzing and evaluating the robustness of the deep generative models based on Mean Absolute Error (MAE), Mean Squared Error (MSE) and Root Mean Squared Error (RMSE) and comparing the statistical results of the synthesized population for the multiple surveys. Again, we discuss the confidence interval estimation of each model trained on specific dataset.

First, we compare the robustness of the two generative models with respect to changes in sample size. Fig.4 and Fig.5 show the results of the CTGAN and VAE respectively at various sample proportion. The results observed from the plot show that the two models are robust under changes in sample proportion. It is evident from the plot that the results obtained for each of the selected samples are relatively similar. However, it is observed that error rate decreases marginally when the sample size increases. This observation validates a study by Subbaswamy et al. (2021) on the evaluation of model robustness and stability to dataset shift where model performance worsens with decrease in sample proportion. Evidently, a decrease in sample size across all the surveys in each model result in increase in error rate. For instance, in Montreal OD survey 2008, the error rate for MAE, MSE and RMSE marginally reduces with percentage increase in the sample size of the train sets. From a sample size of 25% to 100%, we observe a marginal decline in the error rate for MAE, MSE and RMSE. Table 2 shows the results of the metrics for each model at varying sample sizes. It can be observed in the CTGAN model that train set of 25% which produces MAE of 0.2826 marginally decreases to 0.2692 when the model is trained with 100% of the train data. In terms of both the MSE and RMSE, the model performs similarly to the result obtained for the MAE indicating a marginal decline in error rate when sample size increases. This trend occurs in Montreal OD survey of 2013 and 2018. Similarly, the error rate of MAE, MSE and RMSE for the Tabular VAE model decreases with an increase in sample size. In Montreal OD survey 2008, the error rate of MAE is 0.4883 when trained with 25% of sample size. With a train sample size of 100%, the error rate of MAE marginally declines to 0.4133. Again, the MSE and RMSE of 0.523 and 0.8281 respectively reduces marginally to 0.3875 and 0.6982 respectively when the train sample sizes are increased. In Montreal OD survey of 2013 and 2018, both models produce similar prediction of marginal decrease in error rate when there is increase in sample size.

We further extend our analysis by comparing the performance of the models trained on the multiple travel surveys. As shown in Table 2, the performance of the Tabular CTGAN is compared to the Tabular VAE model for the three observations. Even though increase in sample size slightly decreases the error rates, it is also evident from the results that using a few train sets to train each of the models can achieve results that are on par with the case of using the entire train sets. Thus, for Montreal OD survey 2008, using the entire datasets, Tabular CTGAN achieve MAE, MSE and RMSE of 0.2826, 0.2168 and 0.463 respectively, whereas MAE, MSE and RMSE of 0.2692, 0.2076 and 0.4556 respectively are achieved when the 100% of the train sets are used. Similarly, Tabular VAE achieve MAE, MSE and RMSE of 0.4883, 0.523 and 0.8281 respectively when train set of 25% is used whereas MAE, MSE and RMSE of 0.4133, 0.3875 and 0.6982 respectively are obtained when the entire train datasets are used. Similar trend is observed in Montreal OD surveys of 2013 and 2018 which show results that are relatively the same considering the percentage of sample size for model training. It is evident from the results that the deep generative models perform equally good with any amount of datasets from the Montreal OD surveys of 2008, 2013 and 2018.

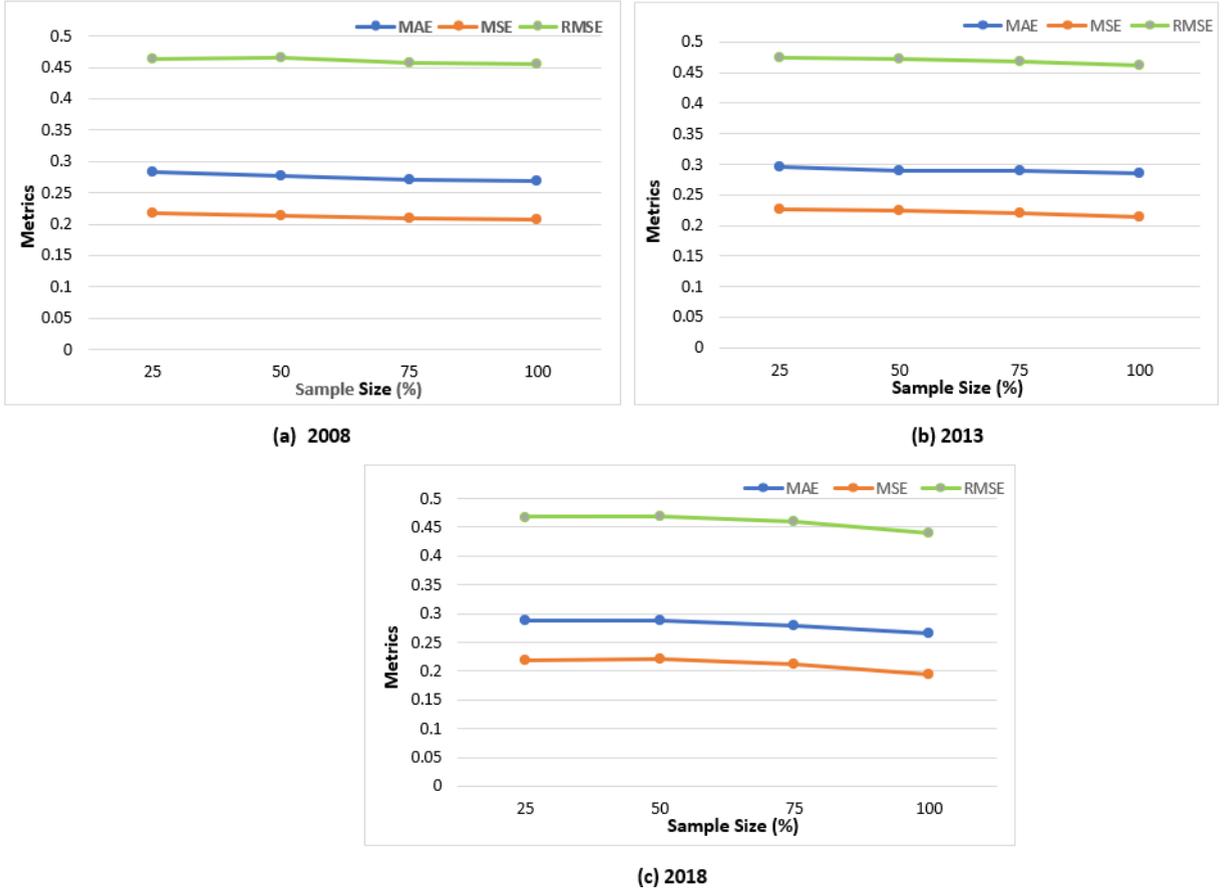

Fig. 4. Error Metrics for GAN for 2008,2013 and 2018 OD Survey

Again, analyzing the performance of each model on the various observations, we observe that the Tabular CT-GAN outperforms the Tabular VAE in terms of MAE, MSE and RMSE. While both models give a good synthetic representation of true data distribution, CTGAN exhibit relatively MAE, MSE and RMSE for all the datasets when compared to the VAE model. The CTGAN achieve a better result in all the observations at different sample size for every performance estimate be it MAE, MSE or RMSE. The loss in approximation of the VAE could be attributed to the low latent dimensional representation adopted by the VAE. Thus, there is a loss of resolution in the synthetic reconstruction, and this validates the study by Badu-Marfo et al. (2020). We detail the performance of the models in Fig. 6 where we plot the bar charts that compare the performance estimates of CTGAN and VAE at different surveys. It is observed that the CTGAN performs better than VAE at every observation based on the error metrics.

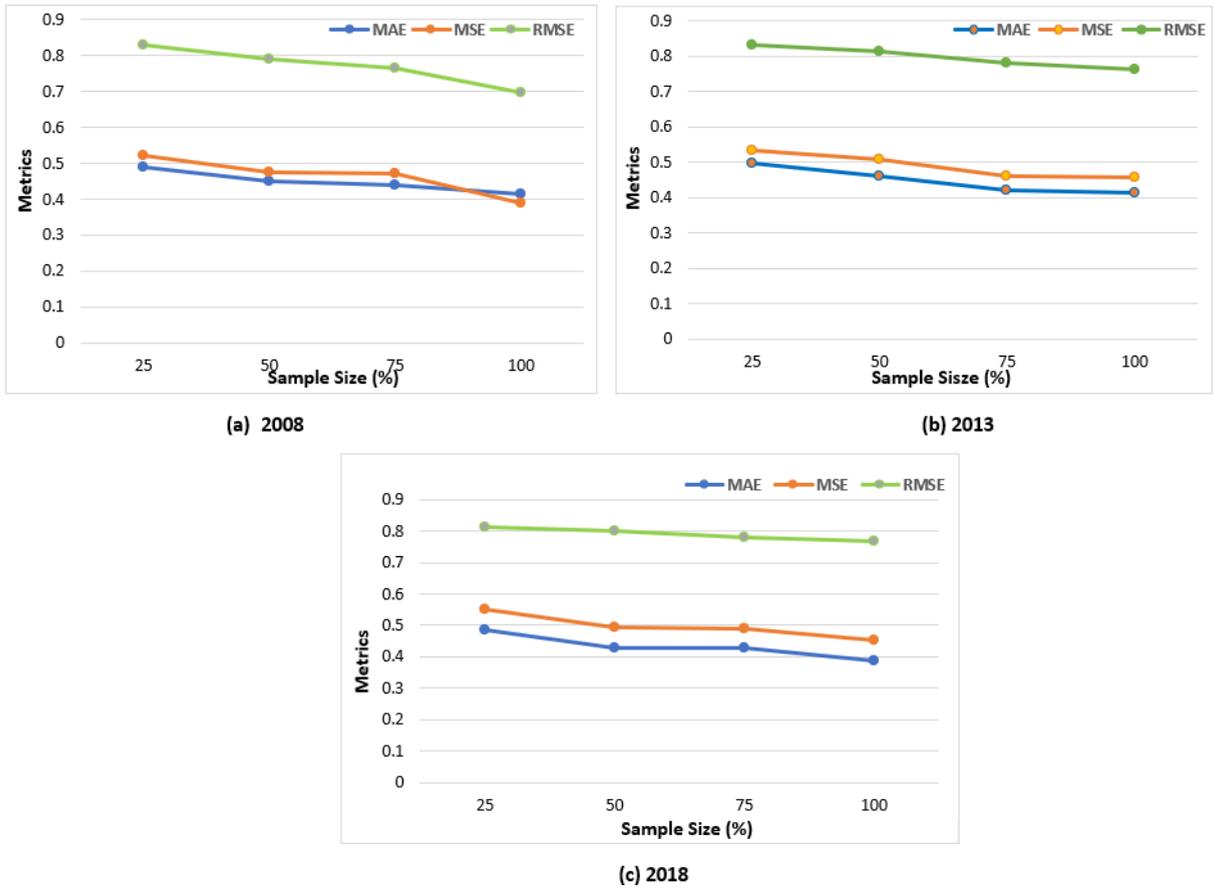

Fig. 5. Error Metrics for VAE for 2008, 2013 and 2018 OD Survey

Table 2. Results for CTGAN and VAE obtained from varying sample sizes of multiple travel surveys

| 2008 Survey | CTGAN | | | VAE | | |
|---|---|---|---|---|---|---|
| Train data | MAE | MSE | RMSE | MAE | MSE | RMSE |
| 25% | 0.2826 | 0.2168 | 0.463 | 0.4883 | 0.5230 | 0.8281 |
| 50% | 0.2763 | 0.2144 | 0.4653 | 0.4510 | 0.4752 | 0.7912 |
| 75% | 0.2699 | 0.2092 | 0.4573 | 0.4381 | 0.4713 | 0.7632 |
| 100% | 0.2692 | 0.2076 | 0.4556 | 0.4133 | 0.3875 | 0.6982 |
| 2013 Survey | CTGAN | | | VAE | | |
| Train data | MAE | MSE | RMSE | MAE | MSE | RMSE |
| 25% | 0.2956 | 0.2259 | 0.4751 | 0.4991 | 0.5345 | 0.8306 |
| 50% | 0.2905 | 0.2236 | 0.4729 | 0.4612 | 0.5069 | 0.8127 |
| 75% | 0.2890 | 0.2201 | 0.469 | 0.4209 | 0.4627 | 0.7825 |
| 100% | 0.2859 | 0.214025 | 0.4626 | 0.4147 | 0.4575 | 0.7616 |
| 2018 Survey | CTGAN | | | VAE | | |
| Train data | MAE | MSE | RMSE | MAE | MSE | RMSE |
| 25% | 0.2876 | 0.22015 | 0.4691 | 0.4847 | 0.5507 | 0.8131 |
| 50% | 0.2874 | 0.2190 | 0.4679 | 0.4295 | 0.4942 | 0.8002 |
| 75% | 0.2782 | 0.2118 | 0.4602 | 0.04267 | 0.4907 | 0.7821 |
| 100% | 0.2651 | 0.1935 | 0.4396 | 0.3861 | 0.4515 | 0.7681 |

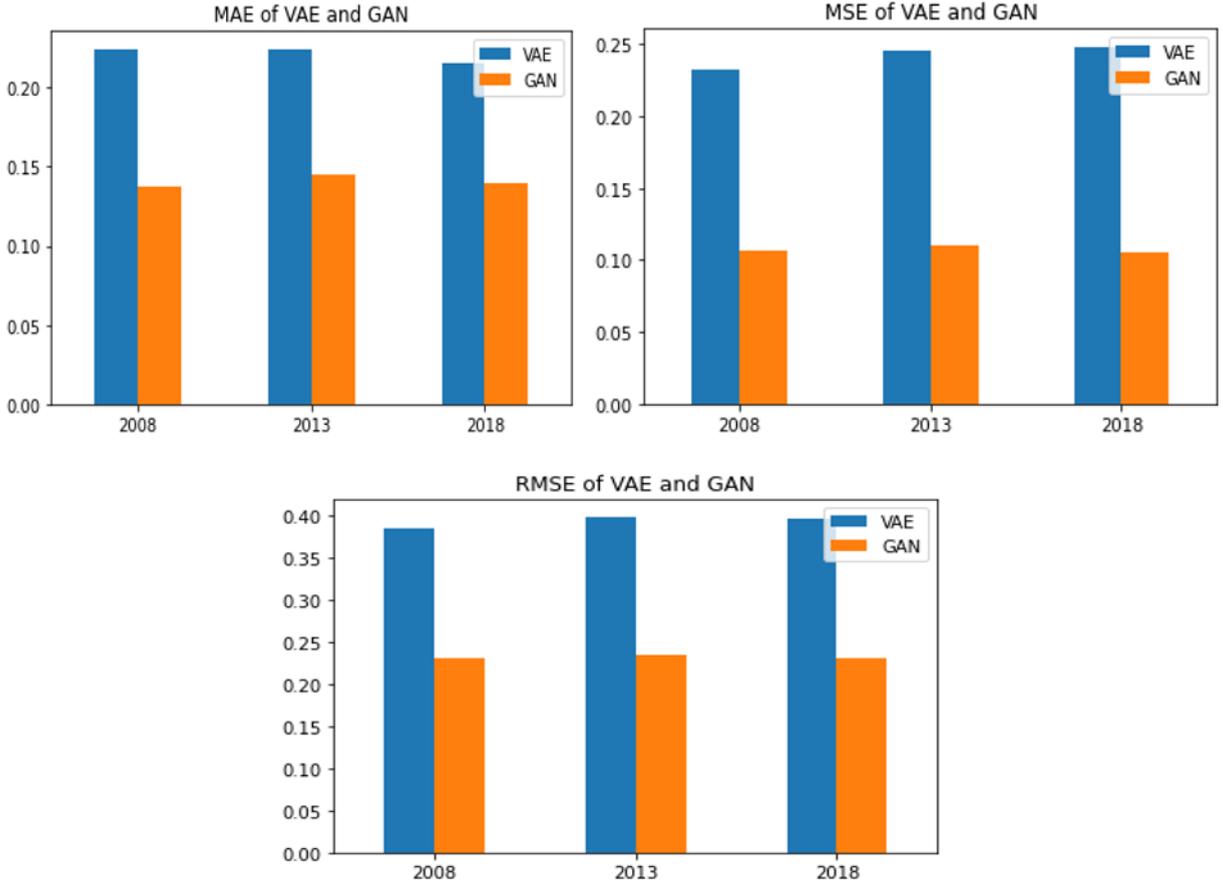

Fig. 6. Comparison of Error Metrics for VAE and GAN at different surveys

## 5.3. Confidence Interval Estimation for Model Stability Analysis

To further extend the analysis, we estimate the confidence intervals of MAE, MSE and RMSE for each model trained on each observation. As a way of quantifying the uncertainty of an estimate, the confidence intervals are determined to estimate the models' performances on unseen data. In our analysis we obtain the confidence interval of each metric based on 16 populations generated from each observation. To obtain the confidence interval, the mean of each error metric is estimated from the range of the 16 populations generated during the model training of each dataset. At 95% confidence interval, we obtain the lower and upper bound of each error metric.

We evaluate the robustness of the deep generative models with respect to the confidence interval obtained from each of the estimated error of the models for each survey. As shown in Table 3, Montreal OD survey 2008 produces a mean of 0.2238 for MAE in the VAE. The prediction is 95% confident that the error is between 0.2114 and 0.2379. The experimental result means that at a repeated experiment, the mean of MAE is expected to be between 21% and 24% at 95% of the time. Again, with a mean of 0.2245 for MAE in the Tabular VAE, the prediction is 95% confident that the error is between 0.2089 and 0.02401 for Montreal OD survey 2013 whereas the mean of 0.2159 for MAE predicts 95% confident interval that the error is between 0.1985 and 0.2351 in Montreal OD 2018. The efficiency of the confidence intervals for the mean of the error estimates of each model in all the surveys is almost same since the data follows a normal distribution as shown in the results from Table 3. It is evident from the results shown in Table 3 that the margin of error between the lower and upper bound at each error metric in each model is minimal for all the observations. This implies that there is 95% likelihood that the range of errors from lower to upper bound covers the true error of each model.

We extend our analysis on model robustness by comparing the confidence intervals of each model at each observation. It is evident from the results that the mean of the error estimates of VAE has a wider variation in the lower and upper bounds when compared to CTGAN. Even though both models show robustness in terms of the mean, the lower and upper bounds of the means of all the error estimates in VAE show wide variations. A wider variation of the confidence interval indicate that VAE is less stable than CTGAN.

Table 3. Results of Confidence Interval for CTGAN and VAE obtained from multiple travel surveys

| | 95% Confidence Interval for Mean of VAE | | | 95% Confidence Interval for Mean of GAN | | |
|---|---|---|---|---|---|---|
| Survey | MAE | Lower | Upper | MAE | Lower | Upper |
| 2008 | 0.2238 | 0.2114 | 0.2379 | 0.1372 | 0.1342 | 0.1404 |
| 2013 | 0.2245 | 0.2089 | 0.2401 | 0.1451 | 0.1432 | 0.1472 |
| 2018 | 0.2159 | 0.1985 | 0.2351 | 0.1398 | 0.1368 | 0.1428 |
| Survey | MSE | Lower | Upper | MSE | Lower | Upper |
| 2008 | 0.2321 | 0.2047 | 0.25503 | 0.1060 | 0.1037 | 0.1084 |
| 2013 | 0.2452 | 0.2301 | 0.2603 | 0.1104 | 0.1080 | 0.1130 |
| 2018 | 0.2484 | 0.2311 | 0.2679 | 0.1055 | 0.1022 | 0.1089 |
| Survey | RMSE | Lower | Upper | RMSE | Lower | Upper |
| 2008 | 0.3851 | 0.3607 | 0.4059 | 0.2301 | 0.2277 | 0.2327 |
| 2013 | 0.3984 | 0.3860 | 0.410 | 0.2349 | 0.2323 | 0.2376 |
| 2018 | 0.3954 | 0.3876 | 0.4033 | 0.2296 | 0.2259 | 0.2333 |

## 6. Discussion and Conclusion

This paper presents an operational approach to analyse the robustness of deep generative models to changes in varying populations. In this study, we adopt the tabular-based Composite Travel GAN (CTGAN) that outputs selected tabular attributes and the Tabular Variational Autoencoder. We compare similarity in statistical information of the synthesized results of the tabular-based CTGAN and the VAE based on activity diaries of three different surveys. Thus, the model is trained on varying sample population from Montreal origin-destination survey in 2008, 2013, and 2018. The CTGAN outperforms the VAE in all the observations with minimum error rates. We evaluate the robustness of the deep generative models based on variation in sample population. Even though an incremental change in sample population slightly decreases the error rate, it cannot be the basis for performance of the model. Subsequently, we use bootstrapping method to estimate the confidence intervals of the models by determining the mean of the error metrics from more populations. We demonstrate that our model can generate synthesized data on different datasets at varying sample sizes while maintaining minimal error.

Robust evaluation of deep generative models is essential in ML research because it enables significant comparisons between different models in different studies. In this study, we define robustness based on the ability of a model to effectively perform and marginally change when trained with different datasets with varying sample sizes. The MAE, MSE and RMSE for each model marginally decreases when there is increase in sample sizes. Mostly, the greater statistical power for recognising patterns in data is based on larger datasets (Raudys et al., 1991). Even though the errors marginally decrease with increase in sample sizes, the model performance cannot be based on that in this study. Another evaluation in this study is the comparison of the deep generative models. The CTGAN model show more robustness than VAE in all the observations with respect to the performance estimates. The MAE, MSE and RMSE of the CTGAN are smaller than the VAE model. A probable reason for the better performance of the CTGAN is its ability to produce points that are very close to the true distribution's support than the VAE. While VAE models generate data close to input data points in Euclidean distance with the shape of the data similar to the true data, they show small deviations in values Mi et al. (2018). On the other hand, GAN distance measures fake point's distance to the manifold, and equals zero as long as the point is on the manifold. To analyze the robustness of the model to changes in sample proportion, we estimate the error estimates of each model for every observation. Even though the error rates decrease with an increase in sample sizes, the models are stable to use under wide range of varying sample sizes and similar unseen data. The comparable errors(MAE, MSE, RMSE) predicted on the test set demonstrates that CTGAN performed on par with the VAE. The results thus indicate that the deep generative models that we used for population synthesis in this study are robust to model the multiple travel diaries.

Moreover, the adoption of confidence interval to compare the performance of the deep generative models indicate that the models are robust with respect to multiple datasets of varying sample sizes. Even though the mean of the error estimates of VAE has a smaller variation in the confidence intervals when compared to CTGAN, the two models have means of the error estimates that are minimal. A crucial aspect of adopting bootstrap confidence interval for robust evaluation of the models is the efficiency of the confidence intervals. A narrow confidence interval shows that the model is robust to changes in datasets. Even though a wider confidence interval may be theoretically valid, such wide error bars can be a constraint to reliability of a deep generative model to population synthesis. The narrower confidence intervals of the error estimates of CTGAN and VAE shows that they can give more precise estimate which indicate their robustness with respect to multiple datasets.

While this paper to the best of our knowledge is the first study that investigates the robustness of deep generative models for population synthesis with respect to multiple datasets, we have identified several interesting new research directions that will be explored in future. First, it will be interesting to evaluate other versions of GAN and VAE for population synthesis. Even though most of the deep generative models are used for image data generation, it will be relevant to explore their use for population synthesis, particularly evaluate their robustness on multiple datasets. Second, we will consider the conventional statistical techniques such as IPF and Gibbs Sampler and evaluate their robustness to population synthesis while comparing their performances with the deep generative models. Finally, future studies will consider privacy concerns by integrating federated machine learning with a deep generative model as a deep neural network architecture for activity diary synthesis and evaluating the robustness of the model against multiple datasets.

**Acknowledgements**

We are grateful to Exo, officially known as Reseau de transport metropolitain, the public transport system provider in Greater Montreal Area, for providing the data for analysis and providing partial funding for this project. We are also grateful to the Canada Research Chair (CRC) Program and Natural Sciences and Engineering Research Council (NSERC) for providing the funding support.